\documentclass[runningheads]{llncs}
\usepackage{graphicx}
\usepackage{amsmath,amssymb} 
\usepackage{color}
\DeclareUnicodeCharacter{2212}{\pm}
\begin{document}
\pagestyle{headings}
\mainmatter

\title{A Global to Local Double Embedding Method for Multi-person Pose Estimation} 
\titlerunning{Double Embedding}
%
\renewcommand{\thefootnote}{0}
\author{Yiming Xu\inst{1} \and
Jiaxin Li\inst{2} \and
Yiheng Peng\inst{3} \and
Yan Ding\inst{2*} \and
Hualiang Wei\inst{4}}
\authorrunning{Yiming Xu, Jiaxin Li et al.}
%
\institute{Yingcai Honors College, University of Electronic Science and Technology of China, Chengdu, China \and
Key Laboratory of Dynamics and Control of Flight Vehicle, Ministry of Education, School of Aerospace Engineering, Beijing Institute of Technology, Beijing 100081, China
\\
\and
School of Computer Science and Technology, Donghua University, Shanghai, China
\and
Department of Automatic Control and Systems Engineering, University of Sheffield, Sheffield S1 3JD, UK
\footnotetext[0]
{
*First two authors equally contributed to this work. Yan Ding is the
corresponding author: dingyan@bit.edu.cn
}
}
\maketitle
\begin{abstract}
Multi-person pose estimation is a fundamental and challenging problem to many computer vision tasks.
Most existing methods can be broadly categorized into two classes: top-down and bottom-up methods.
Both of the two types of methods involve two stages, namely, person detection and joints detection. Conventionally, the two stages are implemented separately without considering their 
interactions between them, and this may inevitably cause some issue intrinsically.
In this paper, we present a novel method to simplify the pipeline by implementing person detection and joints detection simultaneously. 
We propose a Double Embedding (DE) method to complete the multi-person pose estimation task in a global-to-local way. 
DE consists of Global Embedding (GE) and Local Embedding (LE). GE encodes different person instances and processes information covering the whole image and LE encodes the local limbs information. 
GE functions for the person detection in top-down strategy while LE connects the rest joints sequentially which functions for joint grouping and information processing in A bottom-up strategy.
Based on LE, we design the Mutual Refine Machine (MRM) to reduce the prediction difficulty in complex scenarios. 
MRM can effectively realize the information communicating between keypoints and further improve the accuracy.
We achieve the competitive results on benchmarks MSCOCO, MPII and CrowdPose, demonstrating the effectiveness and generalization ability of our method.

\end{abstract}

\section{Introduction}
Human pose estimation aims to localize the human facial and body keypoints (e.g., nose, shoulder, knee, etc.) in the image.
It is a fundamental technique for many computer vision tasks such as action recognition \cite{Ch2015P}, human-computer interaction \cite{DBLP:journals/corr/abs-1811-08264,Fang2018Pairwise}, person Re-ID \cite{Qian2017Pose} and so on.

\begin{figure}
  \centering
  \includegraphics[width=80mm]{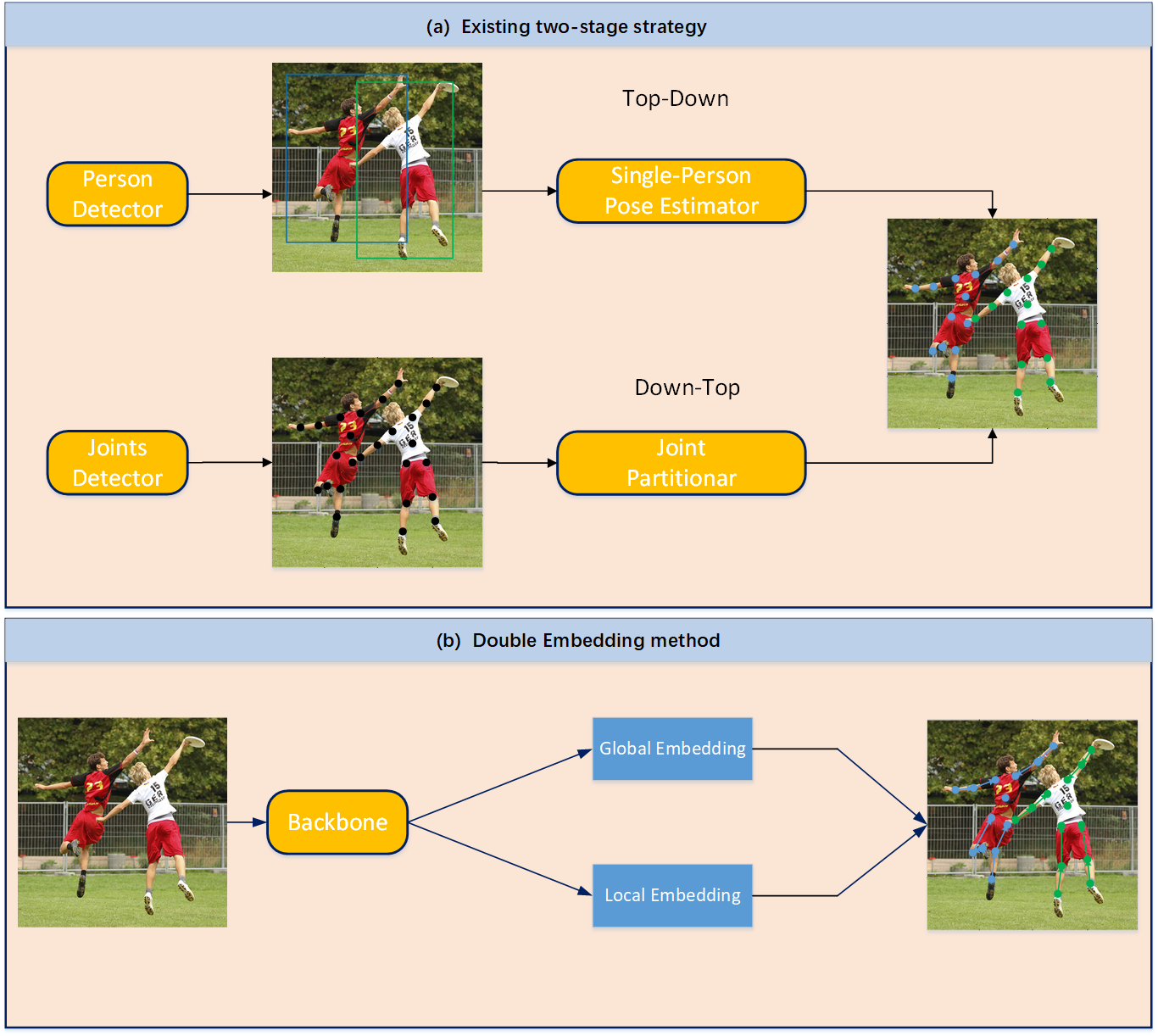} 
  \caption{
    Comparison between (a) existing two-stage strategy and (b) our Double Embedding method for multi-person pose estimation. 
    The proposed DE model implement the person detection and joints detection parallel, overcoming the intrinsic problems of existing two-stage based top-down and bottom-up strategies.}
  \label{fig:1}
  \end{figure}

Most of the existing methods can be broadly categorized into two classes: top-down methods \cite{Xiao2018Simple,Sun2019Deep,Jingdong2020Deep,DBLP:journals/corr/PapandreouZKTTB17,He2018Mask,Huang2017A,Fang2016RMPE,Chen2017Cascaded,Newell2016Stacked} and bottom-up methods \cite{Pishchulin2016DeepCut,Insafutdinov2016,Iqbal2016Multi,osokin2018real,DBLP:journals/corr/NewellD16}.
As shown in Figure 
\ref{fig:1} (a), the top-down strategy first employs a human detector to generate person bounding boxes, and then performs single person pose estimation on each individual person. 
On the contrary, the bottom-up strategy locates all body joints in the image and then groups joints to corresponding persons. 
Top-down strategy is less efficient because the need to perform single person pose estimation on each detected instance sequentially. 
Also, the performance of top-down strategy is highly dependent on the quality of person detections.
Compared to top-down strategy, the complexity of bottom-up strategy is independent of the number of people in the image, which makes it more efficient.
Though as a faster and more likely to be the real-time technique, the bottom-up methods may suffer from solving an NP-hard graph partition problem \cite{Insafutdinov2016,Pishchulin2016DeepCut} to group joints to corresponding persons on densely connected graphs covering the whole image.
\newline

We analyze and try to bypass the disadvantages of these two conventional strategies.
The low efficiency of the top-down strategy comes from the independent single person pose estimation on each person bounding box.
For bottom-up strategy, treating all the detected unidentified joints equally causes the high difficulty of joints grouping process. 
Both top-down and bottom-up strategy are two-stage structure with little interaction between the two stages. 
They both suffer from the separation of person instance detection and joints detection.

\begin{figure}
  \centering
  \includegraphics[width=120mm]{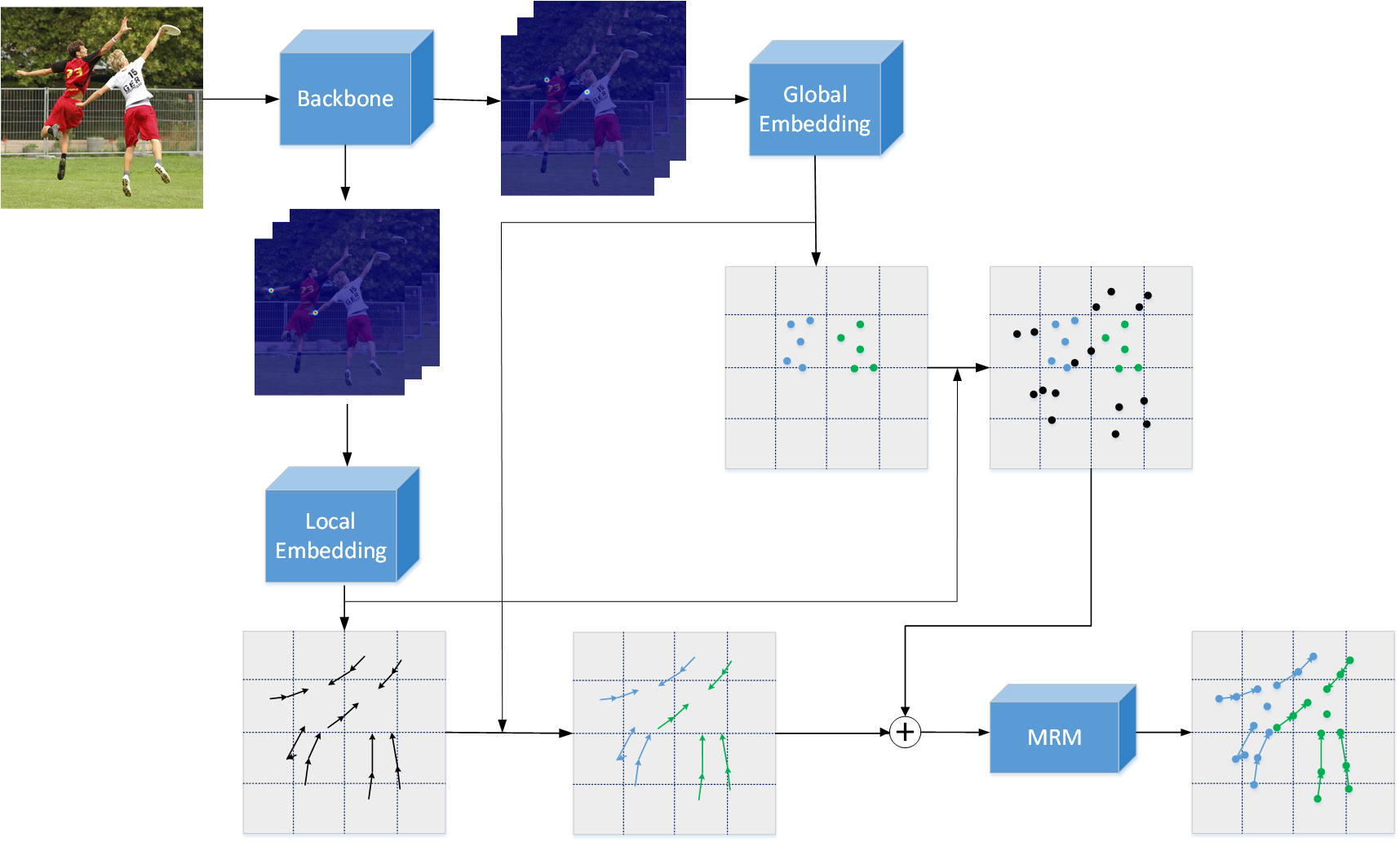} 
  \caption{
    Overview of the Double Embedding (DE) model. For an image, we generate two kinds of feature maps for Global Embedding (GE) and Local Embedding (LE).
    GE and LE works parallel with information communicating to support each other. 
    Based on GE and LE, we design Mutual Refine Machine (MRM) to refine the low-quality predicted joints which further improves the accuracy.
    }
  \label{fig:2}
  \end{figure}

To overcome this intrinsic limitation, we propose to implement person detection and joints detection simultaneously, and realize information communicating between the two procedures to better utilize the structure information of human body. The proposed approach is illustrated in Figure \ref{fig:1} (b).
We observe that the torso joints (shoulders and hips) are more stable than other limbs joints.
With much lower degree of freedom than limbs joints, torso joints can represent the identity information to distinguish the different human instances well.
We also introduce the center joint of the human body.
Center joint is calculated by the average location of annotated joints.
Together with the center joint point, torso joints and center points compose the Root Joints Group (RJG). 
Based on this, we categorize the rest joints on limbs into Adjacency Joints Group (AJG).
In this paper, we propose the Double Embedding method to simplify the pipeline and improve the joints localization accuracy.
The framework of the DE approach is shown in Figure \ref{fig:2}. Double Embedding consists of Global Embedding (GE) and Local Embedding (LE).
GE functions for the person instance separation process through encoding the clustering information of RJG.
We follow the associative embedding method \cite{DBLP:journals/corr/NewellD16} to allocate 1D tags to each pixel related to the joints in root joints group. 
Joints belong to the same person have similar tags while joints belong to different instances have dissimilar tags.

GE encodes global information around the whole image, while LE focuses on local information of each instance based on the global clues from GE. AGJ is connected to identified RJG by corresponding displacement vector field encoded in LE.
Basically, we take the center joint as the reference point.
However, the displacements from extremities joints (wrists and ankles) to center joint are long-range displacements which are vulnerable to background noises. 
To optimize the long-range displacement prediction \cite{DBLP:journals/corr/abs-1803-08225}, we further divide AGJ into two hierarchies: the first level consists of elbows and knees, the second level consists of wrists and ankles. 
AJG is connected sequentially from the second level to first level and finally to torso joints in RJG which are identified. 
Long-range displacements are factorized into accumulative short-range displacements targeting on torso joints (hip joints and shoulder joints). 
Take the left ankle for example, the displacement from ankle to center joint is long-range displacement which is difficult to predict.
To tackle this problem with better utilizing the body structure information, we change the reference joint to left hip, and divide the displacement from left ankle to left hip into shorter displacements: the displacement from left ankle to left knee and the displacement from left knee joint to left hip joint.
Thus, AJG (limbs joints) are connected to RJG and identified in sequence. 
As for facial joints (e.g., Eyes, nose, etc.), we localize them from predicted heatmaps and connect them with the long-range displacements targeting on the center joint.

In addition, we design Mutual Refine Machine (MRM) to further improve the joints localization accuracy and reduce the prediction difficulty in complex scenarios such as pose deformation, cluttered background, occlusion, person overlapping and scale variations. 
Based on hierarchical displacements and connection information encoded in LE, MRP refines the poor-quality predicted joints by high-quality predicted neighboring joints.

We reduce the person detection task to identifying RJG with associative embedding. This is essential to implement the person detection and joints detection at the same time.
This is essential to implement the person detection and the following joints detection and grouping at the same time.
Avoiding the independent single person pose estimation on each detected person bounding boxes makes the method more efficient.
Compared to directly processing all the unidentified joints around the whole image, LE performs local inference with robust global affinity cues encoded in GE, reducing complexity for joints identifying.
Different with the independence of two stages in previous two-stage strategy, GE and LE works mutually to complete the person detection and joints detection parallel.

We implement DE with Convolutional Neural Networks (CNNs) based on the state-of-the-art HigherHRNet\cite{2020HigherHRNet} architecture.
Experiments on benchmarks MSCOCO \cite{Lin2014Microsoft}, MPII \cite{Andriluka2014Human} and CrowdPose \cite{Li2018CrowdPose} demonstrate the effectiveness of our method.

The main contributions of the paper is summarized as follows:
\begin{itemize}
  \item We attempt to simplify the pipeline for multi-person pose estimation, solving the task in global-to-local way.
  \item We propose the Double Embedding method to implement person detection and joints detection parallel, overcoming the intrinsic disadvantages caused by two-stage structure.
  \item Our model achieves competitive performance on multiple benchmarks.
\end{itemize}

\section{Related Works}
\textbf{Top-Down Multi-Person Estimation} Top-down methods \cite{Xiao2018Simple,Sun2019Deep,Jingdong2020Deep,DBLP:journals/corr/PapandreouZKTTB17,He2018Mask,Huang2017A,Fang2016RMPE,Chen2017Cascaded,Newell2016Stacked} first employ object detection \cite{Ren2015Faster,Lin2016Feature,DBLP:journals/corr/abs-1803-06799,Cheng2018Decoupled} to generate person instances within person bounding boxes, and then detect the keypoints of each person by single person pose estimation independently.

Mask R-CNN \cite{He2018Mask} adopts a branch for keypoints detection based on Faster R-CNN \cite{Ren2015Faster}.
G-RMI \cite{DBLP:journals/corr/PapandreouZKTTB17} directly divides top-down methods as two stages and employs independent models for person detection and pose estimation. 
In \cite{Gkioxari2014Using}, Gkioxari et al. adopted the Generalized Hough Transform framework for person instance detection, and then classify joint candidates based on the poselets.
In \cite{Min2011Articulated}, Sun et al. proposed a part-based model to jointly detect person instances and generate pose estimation. 
Recently, both person detection and single person pose estimation benefit a lot from the thriving of deep learning techniques. 
Iqbal and Gall \cite{Iqbal2016Multi} adopted Faster-RCNN \cite{Ren2015Faster} for person detection and convolutional pose machine \cite{7780880} for joints detection.
In \cite{fang2017rmpe}, Fang et al. used spatial transformer network \cite{Jaderberg2015Spatial} and Hourglass network \cite{Newell2016Stacked} for joints detection.
\newline
Though these methods have achieved excellent performance, they suffer from high time complexity due to sequential single person pose estimation on each person proposal. 
Differently, DE performs the person detection and joints detection parallel, which simplifies the pipeline.
\newline
\textbf{Bottom-Up Multi-Person Pose Estimation} Bottom-up methods \cite{Pishchulin2016DeepCut,Insafutdinov2016,Iqbal2016Multi,osokin2018real,DBLP:journals/corr/NewellD16} detect all the unidentified joints in an image and then group them to corresponding person instances. 
\newline
Openpose \cite{DBLP:journals/corr/CaoSWS16} proposes part affinity field to represent the limbs. The method calculates line integral through limbs and connects joints with the largest integral.
In \cite{DBLP:journals/corr/NewellD16}, Newell et al. proposed associate embedding to assign each joint with a 1D tag and then group joints which have the similar tags. 
PersonLab \cite{DBLP:journals/corr/abs-1803-08225} groups joints by a 2D vector field in the whole image. In \cite{DBLP:journals/corr/abs-1903-06593}, Sven Kreiss et al. proposed Part Intensity Field (PIF) to localize body parts and Part Association Field (PAF) to connect body parts to form full human poses.
\newline
Nevertheless, the joints grouping cues of all these methods are covering the whole image, which causes high complexity for joints grouping. 
Different with the prior methods, global clues from GE reduce the search space for graph partition problem, avoiding high complexity of joint partition in bottom-up strategy.
\newline
\section{Double Embedding Method}
In this section, we present our proposed Double Embedding method. Figure \ref{fig:2} illustrates the overall architecture of the proposed approach.
\newline
\textbf{Joints Feature Map} For an image \textbf{\emph{I}}, we generate two kinds of feature maps from backbone network, one for Global Embedding (GE) and one for Local Embedding (LE).
We use $J_R=\{J_{R1},J_{R2},...,J_{RU}\}$ to denote Root Joints Group (RJG), $J_{Ri}$ is the i-th kind of joints in RJG for all N persons in image \textbf{\emph{I}},
and \emph{U} is the number of joint categories in RJG. Similarly, we use $J_A=\{J_{A1},J_{A2},...,J_{AV}\}$ to denote Adjacency Joints Group (AJG), $J_{Ai}$ is the i-th kind of joints in AJG for all persons in image \textbf{\emph{I}}, and
{$V$} is the number of joint categories in RJG. For Global Embedding, let $h_{Gk}\in{R^{W\times{H}}}(k=1,2,...,U)$
denote the feature map for the k-th kind of joint in Root Joints Group.
For Local Embedding, $h_{Lk}\in{R^{W\times{H}}}(k=1,2,...,V)$ denotes the feature map for the k-th kind of joint in AJG.
The form and generation method of $h_{Gk}$ and $h_{Lk}$ are the same. To simplify
the description, we use $h_{fk}$ to represent both $h_{Gk}$ and $h_{Lk}$. 

It was pointed out that Directly regressing the absolute joint coordinates in an image is difficult\cite{DBLP:journals/corr/CarreiraAFM15,Pishchulin2016DeepCut}. We therefore use heatmap, a confidence map models the joints position as Gaussian peaks. 
For a position $(x,y)$ in image \textbf{\emph{I}}, $h_{fk}(x,y)$ is calculated by:
\begin{equation}
h_{fk}(x,y)=
\begin{cases}
exp(-\frac{\left \| (x,y)-(x_k^i,y_k^i) \right \|_2^2}{\sigma^2}), &(x,y)\in{\aleph_k^i} \\
0, &\text{otherwhise}
\end{cases}
\end{equation}
In which $\sigma$ is an empirical constant to control the variance of Gaussian distribution, set as 
7 in our experiments. $(x_k^i,y_k^i)$ denotes the i-th groundtruth joint position in $h_{fk}$. $\aleph_k^i={(x,y) \big | \left \| (x,y)-(x_k^i,y_k^i) \right \|_2 \leq \tau}$
is regressing area for each joint to truncate the Gaussian distribution. Thus, we generate two kinds of feature map for GE and LE.
Joints location of RJG and AJG are derived through NMS process.
\newline
\textbf{Global Embedding} Global Embedding functions as a simpler person detection process, reduces person detection problem to identifying the RJG. 
We use Associate Embedding for this process. 
The identification information in $h_{Gk}$ is encoded in $1D$ tag space $T_k$ for k-th
joint in RJG. Based on the pixel location derived from peak detections in feature map $h_{Gk}$, 
corresponding tags are retrieved at the same pixel location in $T_k$. 
Joints belonging to one person have similar tags while tags of joints in different persons have obvious difference.
Let $p=\{p_1,p_2,...,p_N\}$ denote the $N$ persons containing in image \textbf{\emph{I}}.
GE can be represented as a function $f_{GE}:h_{Gk}\rightarrow T_k$. $f_{GE}$ learns to densely transform every pixel in $h_{Gk}$ to embedding space $T_k$.
We use $loc_{(p_n,J_{Rk})}(n=1,2,...,N,k=1,2,...,U)$ to denote the ground truth pixel location of the k-th kind of joint in RJG of n-th person.

If $U^{'}$ joints are labeled, the reference embedding for the n-th person is the average of retrieved tags of RJG in this person:
\begin{equation}
\bar{Tag}_n={\frac{1}{U^{'}}}\sum_{k} T_k(loc_{(p_n,J_Rk)}) 
\end{equation}
To pull the tags of joints within an individual together, pull-loss computes the squared distance between the reference embedding and the predicted embedding for each joint.
\begin{equation}
L_{pull}={\frac{1}{NU^{'}}}\sum_n\sum_k(\bar{Tag}_n-T_k(loc_{(p_n,J_{Rk})}))^2
\end{equation}
To push the tags of joints in different persons, push-loss penalizes the reference embeddings that are close to each other. As the distance between two tags increases, push-loss drops exponentially to zero resembling probability density function of Gaussian distribution:
\begin{equation}
L_{push} = {\frac{1}{N^2}}{\sum_n\sum_{n^{'}}}exp \{ -{\frac{1}{2\sigma^2}}(\bar{Tag}_n-{\bar{Tag}_{n^{'}}} )^2 \}
\end{equation}
The loss to train the model as $f_{GE}$ is the sum of $L_{pull}$ and $L_{push}$:
\begin{equation}
L^G={\frac{1}{NU^{'}}}\sum_n\sum_k(\bar{Tag}_n-T_k(loc_{(p_n,J_{Rk})}))^2 + {\frac{1}{N^2}}{\sum_n\sum_{n^{'}}}exp \{ -{\frac{1}{2\sigma^2}}(\bar{Tag}_n-{\bar{Tag}_{n^{'}}} )^2 \}
\end{equation}
\textbf{Local Embedding} Local Embedding performs local inference and builds the connection clues between AJG and identified RJG.
Relative position information of $h_{Lk}$ is encoded in displacement space $D_{k}$ for k-th joint in AJG. 
For each joint in AJG, we get its corresponding normed displacement to RJG from $D_k$ at the same position in $h_{Lk}$.
We first build the basic displacement to connect to the center joint.
Basic displacement for the k-th joint in AJG of the n-th person is represented as the 2D vector:
\begin{equation}
Dis_n^k=(x_n^r,y_n^r)-(x_n^k,y_n^k)
\end{equation}
In which, $(x_n^r,y_n^r)$ is the location of the center joint in the n-th person.
Besides, we design the hierarchical displacement to connect the limbs joints to corresponding torso joints.
Compared to basic displacements directly targeting on the center joint, hierarchical displacements are shorter ones which are more robust and easier to predict. 
Normally, we use hierarchical displacements in inference process. But if some intermediate joints are absent, we directly use the long-range prediction to complete the inference. 

The general displacement for joint A to joint B of n-th person is:
\begin{equation}
Dis_n^{A2B}=(x_n^B,y_n^B)-(x_n^A,y_n^A)
\end{equation}
In some cases, we may use the property $Dis_n^{B2A}=-Dis_n^{A2B}$ to
get reverse displacements of paired joints.

Local Embedding $f_{LE}:h_{Lk}\rightarrow D_k$ maps each pixel in feature map $h_{Lk}$ to the embedding space $D_k$. For learning $f_{LE}$, we build the target regression map $T_n^A$ 
for the displacement vector from joint A (the k-th kind of joint in AJG) to joint B of the n-th person as follows:
\begin{equation}
D_k^A(x,y)=
\begin{cases}
Dis_n^{(x,y)2(x_n^B,y_n^B)}/Z,& \text{if}(x,y)\in \aleph_k^A \\
0, & \text{otherwise}
\end{cases}
\end{equation}
where $\aleph_k^A=\{ (x,y)|\|(x,y)-(x_n^A,y_n^A) \|_2 \leq \tau\}$.
The displacements are created in $\aleph_k^A$ which is the same as regression area in $h_{Lk}$.
$Z=\sqrt{H^2+W^2}$ is the normalization factor, with H and W denoting the height and width of the image.

The starting point A $(x_n^A,y_n^A)$ is generated from peak detections in feature map $h_{Lk}$, we get 
its corresponding displacement to joint B in $D_k$ as $Dis_n^{A2B}=D_k((x_n^A,y_n^A))$. The ending joint is obtained:
$(x_n^{End},y_n^{End})=(x_n^A,y_n^A)+Z\cdot{Dis_n^{A2B}}$. 
Compared with the peak detections in $h_{LB}$
(containing the same category joints as joint B for all persons in the image, including $(x_n^B,y_n^B)$), 
it will be confirmed that joint B is the ending joint of joint A. Accordingly, joint A is connected to joints B, meaning they share the same identification.
In this way, joints in AJG are connected to RJG and identified.
\newline
\textbf{Mutual Refine Machine} We design the Mutual Refine Machine (MRM) to reduce the prediction difficulty in complex scenes.
For a low-quality predicted joint, it can be refined by the neighboring high-quality joints.
Based on the displacements and connection information in LE, MRM realizes the information communicating between paired joints. 
For n-th person, if prediction probability of i-th joint in k-th kind of joints confidence map $h_{fk}((x_n^i,y_n^i))$ is lower than its neighboring paired joints 
$\{h_{fk^{\prime}}(({x_n^{i^\prime}}, {y_n^{i^\prime}}))\}(\text{in which } h_{fk^{\prime}}((x_n^{i^\prime},y_n^{i^\prime}))>0.75)$, then we refine the location of i-th joint with the weighted fusion of its neighboring joints. Refined location is:
\begin{equation}
  \left( x_n^i,y_n^i \right)_{refined}={\frac{h_{fk}(({x_n^{i}}, {y_n^{i}}))}{Q}}\ast(x_n^i,y_n^i)+\sum_{i^\prime}{\frac{h_{fk^{\prime}}(({x_n^{i^\prime}}, {y_n^{i^\prime}}))}{Q}}\ast((x_n^{i^\prime},y_n^{i^\prime})+Dis_n^{i^\prime2i})
\end{equation}
\begin{equation}
    Q=h_{fk}(({x_n^{i^\prime}}, {y_n^{i^\prime}}))+\sum_{i^\prime}{h_{fk^{\prime}}(({x_n^{i^\prime}}, {y_n^{i^\prime}}))}
\end{equation}
\textbf{Training and inference} To train our model, we adopt L2 loss $L^H$ for joint
confidence regression, smooth L1 loss \cite{Girshick2015Fast} $L^D$ for displacements regression and $L^G$ for GE.
The total loss L for each image is the weighted sum of $L^H$, $L^D$ and $L^G$:
\begin{equation}
  L={\sum_{x=1}^{U+V}L^H(h_{fx},\hat{h}_{fx})}+{\alpha\sum_{y=1}^V L^D(D_y,\hat{D}_y)}+\beta\sum_{z=1}^U L^G
\end{equation}
where $\hat{h}_{fx}$ and $\hat{D}_y$ denote the predicted joints  confidence map and displacements regression map. 
$\alpha$ and $\beta$ are constant weight factor to balance three kinds of losses, both set as 0.01. The overall framework of DE is end-to-end trainable via gradient backpropagation.

The overall architecture of DE is illustrated in Figure \ref{fig:2}. For an image, DE generates two kinds of feature maps ${\hat{h}}_{Gk}$ and ${\hat{h}}_{Lk}$. through performing NMS and on them, we get predicted joints location of RJG and AJG.
GE gives identification tags for RJG and LE provides the connection relation to connect AJG to RGJ. 
To better present the collaborative work of GE and LE, we add the intermediate illustration.
Connected pairs get identification information from GE and joints in GE expand to all joints by connectivity in LE. 
Based on the displacements and connectivity in LE, MRM refines the low-quality predicted joints.
The final result is generated through the combination of refined results from GE and LE.
\section{Experiments}
\subsection{Experiment setup}
\textbf{Datasets} We evaluate the proposed Double Embedding model on three widely used benchmarks for multi-person pose estimation: MSCOCO \cite{Lin2014Microsoft} dataset, MPII \cite{Andriluka2014Human} dataset and CrowdPose \cite{Li2018CrowdPose} dataset.

The MSCOCO \cite{Lin2014Microsoft} dataset contains over 200, 000 images and 250, 000 person instances labeled with 17 keypoints.
COCO is divided into train/val/test-dev sets with 57k, 5k and 20k images respectively.
MPII \cite{Andriluka2014Human} dataset contains 5,602 images of multiple persons.
Each person is annotated with 16 body joints. Images are divided into 3,844 for training and 1,758 for testing.
MPII also provides over 28,000 annotated single-person pose samples.
The CrowdPose\cite{Li2018CrowdPose}  dataset consists of 20,000 images, containing about 80,000 person instances.
The training, validation and testing subset are split in proportional to 5:1:4. CrowdPose has more crowded scenes than the COCO and MPII, and therefore is more challenging for multi-person pose estimation.
\newline
\textbf{Data augmentation} We follow the conventional data augmentation strategies in experiments.
For MSCOCO and CrowdPose datasets, we augment training samples with random rotation $([-30^{\circ},30^{\circ}])$, random scale $[0.75, 1.5]$, random translation $[−40, 40]$ and random horizontally flip to crop input images to 640x640 with padding.
For MPII dataset, random scale is set as ([0.7, 1.3]) while other augmentation parameters are set the same as MSCOCO and CrowdPose datasets.
\newline
\textbf{Evaluation metric} For COCO and CrowdPose datasets, the standard evaluation metric is based on Object Keypoint Similarity (OKS):
\newline
\begin{equation}
  OKS = \frac{\sum_{i}exp(-d_i^2/2s^2k_i^2)\delta(v_i>0)}{\sum_{i}\delta(v_i>0)}
\end{equation}
where $d_i$ is the Euclidean distance between the predicted joints and ground truth,
$v_i$ is the visibility flag of the ground truth, $s$ is the object scale, and $k_i$ is 
a per-keypoint constant that controls falloff. The standard average precision and recall scores are shown as:
$AP^{50}$(AP at OKS = 0.50), $AP^{75}$, AP (the mean of AP scores at 10 positions, OKS = 0.50, 0.55, . . . , 0.90, 0.95;${AP}^M$ for medium objects, ${AP}^L$ for large objects, and AR at OKS = 0.50, 0.55, . . . , 0.90, 0.955.

For MPII dataset, the standard evaluation metric is PCKh (head-normalized probability of correct keypoint) score. 
A joint is correct if it falls within $al$ pixels of the groundtruth position, where
$\alpha$ is a constant and $l$ is the head size that corresponds to  60\% of the diagonal length of the ground-truth head bounding box. The PCKh@0.5 ($\alpha$ = 0.5) score is reported. 
\newline
\textbf{Implementation} For COCO dataset, we use standard validation set for ablation studies while use test-dev set to compare with other state-of-the-art methods. 
or CrowdPose dataset, we use CrowdPose train and val set to train our model, and use test set for validation. For MPII dataset, following \cite{nie2019singlestage}, we randomly select 350 groups of multi-person training samples as the validation dataset and use the remaining training samples and all single-person images as train dataset. 
We use the Adam optimizer \cite{Kingma2014Adam}. For COCO and CrowdPose datasets, the base learning rate is set to 1e-3, and dropped to 1e-4 and 1e-5 at the 200th and 260th epochs respectively. 
We train the model for a total of 300 epochs. For MPII dataset, we initialize learning rate by 1e-3. We train the model for 260 epochs and decrease learning rate by a factor of 2 at the 160th, 180th, 210th, 240th epoch. Following [HigherHRNet-30], we adopt flip test for all the experiments.
\setlength{\tabcolsep}{4pt}
\begin{table}[h]
\begin{center}
\caption{
  Comparison with state-of-the-arts on COCO2017 test-dev dataset. Top: w/o multi-scale test. Bottom: w/ multi-scale test.
}
\label{table:headings1}
\begin{tabular}{llllll}
\hline\noalign{\smallskip}
Method $\qquad\qquad$& $AP$ & $AP^{50}$ & $AP^{75}$ & $AP^M$ & $AP^L$\\
\hline\noalign{\smallskip}
\multicolumn{6}{l}{w/o multi-scale test} \\
\hline\noalign{\smallskip}
CMU-Pose\cite{osokin2018real}$\dagger$  & 61.8 & 84.9 & 67.5 & 57.1 & 68.2\\
RMPE\cite{Fang2016RMPE} & 61.8 & 83.7 & 69.8 & 58.6 & 67.6\\
Associate Embedding\cite{DBLP:journals/corr/NewellD16} & 62.8 & 84.6 & 69.2 & 57.5 & 70.6\\
Mask-RCNN\cite{He2018Mask}$\dagger$ & 63.1 & 87.3 & 68.7 & 57.8 & 71.4 \\
G-RMI\cite{DBLP:journals/corr/PapandreouZKTTB17}$\dagger$ & 64.9 & 85.5 & 71.3 & 62.3 & 70.0\\
PersonLab\cite{DBLP:journals/corr/abs-1803-08225} & 66.5 & 88.0 & 72.6 & 62.4 & 72.3\\
PifPaf\cite{DBLP:journals/corr/abs-1903-06593} & 66.7 & - & - & 62.4 & 72.9\\
SPM\cite{nie2019singlestage} & 66.9 & 88.5 & 72.9 & 62.6 & 73.1\\
HigherHRNet\cite{cheng2019higherhrnet} & 68.4 & 88.2 & 75.1 & 64.4 & 74.2\\
CPN\cite{Chen2017Cascaded}$\dagger$ & 72.1 & 91.4 & 80.0 & 68.7 & 77.2\\
DoubleEmbedding(Ours) & 69.7 & 88.4 & 76.9 & 65.8 & 75.1\\
\hline\noalign{\smallskip}
\multicolumn{6}{l}{w/ multi-scale test} \\
\hline\noalign{\smallskip}
    Hourglass\cite{DBLP:journals/corr/NewellD16}  & 65.5 & 86.8 & 72.3 & 60.6 & 72.6\\
    Associate Embedding\cite{DBLP:journals/corr/NewellD16} & 65.5 & 86.8 & 72.3 & 60.6 & 72.6\\
    PersonLab\cite{DBLP:journals/corr/abs-1803-08225} & 68.7 & 89.0 & 75.4 & 64.1 & 75.5\\
    HigherHRNet\cite{cheng2019higherhrnet} & 70.5 & 89.3 & 77.2 & 66.6 & 75.8\\
    DoubleEmbedding(Ours) & 71.6 & 89.5 & 78.6 & 68.8 & 76.0\\
\hline\noalign{\smallskip}
$\dagger$ indicates\quad top-down\quad methods
\end{tabular}
\end{center}
\end{table}
\setlength{\tabcolsep}{1.4pt}
\subsection{Results on COCO dataset} 
\textbf{Comparison with state-of-the-arts} In Table 1, we compare our proposed model with other state-of-the-arts methods on COCO2017 test-dev dataset. We test the run time of single-scale inference, 
the proposed method realizes the balance on speed and accuracy. 
We achieve the competitive accuracy which outperforms most existing bottom-up methods. 
Compared to the typical top-down method CPN\cite{Chen2017Cascaded}, we narrow the gap to top-down method in accuracy with less complexity. 
This demonstrates the effectiveness of DE on multi-person pose estimation.
\newline
\textbf{Ablation analysis} We conduct ablation analysis on COCO2017\cite{Lin2014Microsoft} validation dataset without multi-scale test.
We evaluate the impact of the introduced hierarchical short-range displacements that factorize the basic displacements.
Also, the effect of MRM is studied. MRM is implemented based on the hierarchical displacements, so MRM is non-existent without hierarchical displacements. 
Results are shown in Table 2. 
which shows that with basic displacements only, DE achieves 69.3\% mAP. By introducing hierarchical displacements, performance improves to 70.7\% mAP with 1.3\% mAP increasing.
Based on the hierarchical displacements, MRM further improve 0.9\% mAP. The result shows the effectiveness of hierarchical displacements and MRM.
\setlength{\tabcolsep}{4pt}
\begin{table}
\begin{center}
\caption{
  Ablation experiments on COCO validation dataset.
}
\label{table:headings2}
\begin{tabular}{llllllll}
\hline\noalign{\smallskip}
\multicolumn{3}{l}{Model Settings} &\multicolumn{5}{l}{Pose Estimation} \\
\hline\noalign{\smallskip}
Basic Dis. & Hierar Dis. & MRM & $AP$ & $AP^{50}$ & $AP^{75}$ & $AP^M$ & $AP^L$\\
\hline\noalign{\smallskip}
\checkmark &  &  & 69.3 & 87.0 & 75.9 & 65.1 & 76.6\\
\checkmark & \checkmark &  & 70.7 & 88.1 & 76.8 & 66.5 & 76.4\\
\checkmark & \checkmark & \checkmark & 71.6 & 88.3 & 77.5 & 66.9 & 77.8\\
\hline
\end{tabular}
\end{center}
\end{table}
\setlength{\tabcolsep}{1.4pt}

\begin{figure}
  \centering
  \includegraphics[width=100mm]{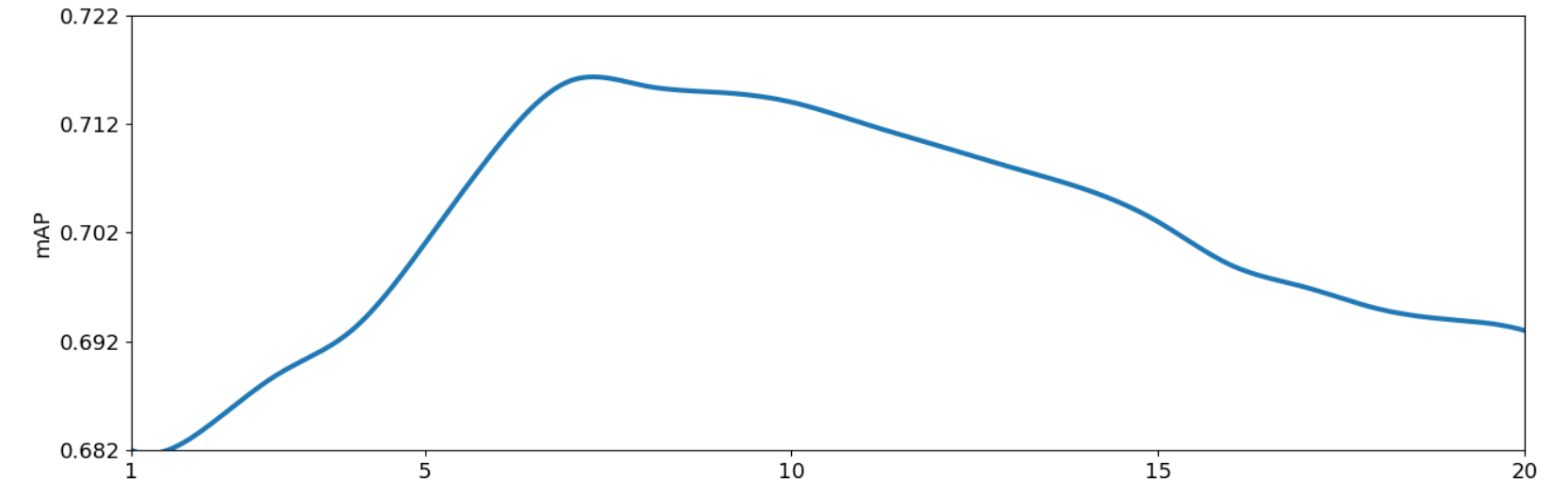} 
  \caption{
    Studies on hyper-parameter $\tau$, which decides the regression area for joints confidence map and displacements.
    }
  \label{fig:3}
\end{figure}

In addition, we analyze the impact of the hyper-parameter $\tau$ which decides the regression area for joints confidence map and displacements in Section 3. 
We observe the performance of proposed model as $\tau$ varies from 1 to 20. 
As shown in Figure \ref{fig:3}, the performance monotonically improves as $\tau$  increases from 1 to 7.
When $7<\tau<10$, performance remains unchanged as $\tau$ increases.
When $\tau>10$, performance drops as $\tau$ increases. 
This can be explained by the distribution of positive samples of the dataset. 
When $\tau$ increases in the range between 1 and 7, positive samples increase and larger effective area of  joints is counted in joints confidence and displacements regression in training.
When $\tau$ increases between 7 and 10, effective information and background noise increases with equivalent effect.
When $\tau>10$, more background noise is countered as positive samples, regression area of joints overlaps with each other as $\tau$ keeps increasing.
Smaller $\tau$ means less complexity, thus we set $\tau=7$ for balancing the accuracy and efficiency.
\newline
\textbf{Qualitative results} Qualitative results on COCO dataset are shown in the top row of Figure \ref{fig:4}. 
The proposed model performs well in challenging scenarios, e.g., pose deformation (1st and 2nd examples), person overlapping and self-occlusion (3rd example), crowded scene (4nd example), and scale variation and small-scale prediction (5st example). 
This presents the effectiveness of our method.
\subsection{Results on MPII dataset}
\setlength{\tabcolsep}{4pt}
\begin{table}
\begin{center}
\caption{
  Comparison with state-of-the-arts on the full testing set of MPII dataset.
}
\label{table:headings3}
\begin{tabular}{llllllllll}
\hline\noalign{\smallskip}
Method $\qquad\qquad$ & Head & Sho & Elb & Wri & Hip & Knee & Ank & Total & Time[s]\\
\noalign{\smallskip}
\hline
\noalign{\smallskip}
lqbal and Gall\cite{Iqbal2016Multi}  &58.4  &53.9  &44.5  &35.0  &42.2  &36.7  &31.1  &43.1  &10\\
Insafutdinov et al.\cite{Insafutdinov2016}  &78.4  &72.5  &60.2  &51.0  &57.2  &52.0  &45.4  &59.5  &485\\
Levinkov et al.\cite{Insafutdinov2016}  &89.8  &85.2  &71.8  &59.6  &71.1  &63.0  &53.5  &70.6  &-\\
Insafutdinov et al.\cite{DBLP:journals/corr/InsafutdinovAPT16}  &88.8  &87.0  &75.9  &64.9  &74.2  &68.8  &60.5  &74.3  &-\\
Cao et al.\cite{DBLP:journals/corr/CaoSWS16}  &91.2  &87.6  &77.7  &66.8  &75.4  &68.9  &61.7  &75.6  &0.6\\
Fang et al.\cite{Fang2016RMPE}  &88.4  &86.5  &78.6  &70.4  &74.4  &73.0  &65.8  &76.7  &0.4\\
Newell and Deng\cite{Newell2016Stacked}  &92.1  &89.3  &78.9  &69.8  &76.2  &71.6  &64.7  &77.5  &0.25\\
Fieraru et al.\cite{DBLP:journals/corr/abs-1804-07909}  &91.8  &89.5  &80.4  &69.6  &77.3  &71.7  &65.5  &78.0  &-\\
SPM\cite{nie2019singlestage}  &89.7  &87.4  &80.4  &72.4  &76.7  &74.9  &68.3  &78.5  &0.058\\
DoubleEmbedding (Ours)  &91.9  &89.7  &81.6  &74.9  &79.8  &75.8  &71.5  &80.7  &0.21\\
\hline
\end{tabular}
\end{center}
\end{table}
\setlength{\tabcolsep}{1.4pt}

\setlength{\tabcolsep}{4pt}
\begin{table}
\begin{center}
\caption{
  Ablation experiments on MPII validation dataset.
}
\label{table:headings4}
\begin{tabular}{llllllllllll}
\hline\noalign{\smallskip}
\multicolumn{3}{l}{Model Settings} &\multicolumn{8}{l}{Pose Estimation} \\
\hline\noalign{\smallskip}
Basic Dis. & Hierar Dis. & MRM & Head & Sho & Elb & Wri & Hip & Knee & Ank & Total\\
\hline\noalign{\smallskip}
\checkmark &  &  &92.1  &88.4  &78.3  &68.9  &77.5  &73.6  &63.7  &77.5\\
\checkmark & \checkmark &  &92.3  &89.2  &79.8  &71.3  &78.1  &74.8  &66.2  &78.8\\
\checkmark & \checkmark & \checkmark &92.3  &90.1  &81.2  &72.6  &79.0  &75.7  &67.1  &79.7\\
\hline
\end{tabular}
\end{center}
\end{table}
\setlength{\tabcolsep}{1.4pt}

Table 3 shows a comparison of the proposed method with state-of-the-arts methods on MPII dataset.
The proposed model obtains 80.7\% mAP achieving competitive result among other bottom-up methods.
In addition, we conduct the ablation study on MPII validation dataset to verify MRM and the hierarchical displacements compared with the basic displacements.
As shown in table 4, DE improves from 77.5\% mAP to 78.8\% mAP by introducing hierarchical displacements.
Moreover, on wrists and ankles are significant from 68.9\% to 71.3\% mAP and 73.6\% to 74.8\% mAP, respectively. Indicating the effectiveness of hierarchical displacements to factorize the long-range displacements.
MRM further improves 1.1\% mAP based on hierarchical displacements. 

Qualitative results on MPII are shown in the middle row of Figure \ref{fig:4}, demonstrating the good performance and robustness of our model in complex scenes such as person scale variations (1st example), large pose deformation (2nd and 3rd examples) and small-scale prediction (3rd example).
\newline
\subsection{Results on CrowdPose dataset}
\setlength{\tabcolsep}{4pt}
\begin{table}
\begin{center}
\caption{
Comparison with state-of-the-arts on CrowdPose test dataset.
}
\label{table:headings5}
\begin{tabular}{lllllll}
\hline\noalign{\smallskip}
Method $\qquad\qquad$& $AP$ & $AP^{50}$ & $AP^{75}$ & $AP^M$ & $AP^L$ & $AP^H$\\
\noalign{\smallskip}
\hline
\noalign{\smallskip}
Openpose\cite{osokin2018real}  &-  &-  &-  &62.7  &48.7  &32.3\\
Mask-RCNN$\dagger$\cite{He2018Mask}  &57.2  &83.5  &60.3  &69.4  &57.9  &45.8\\
AlphaPose$\dagger$\cite{Fang2016RMPE}  &61.0  &81.3  &66.0  &71.2  &61.4  &51.1\\
SPPE$\dagger$\cite{Li2018CrowdPose}  &66.0  &84.2  &71.5  &75.5  &66.3  &57.4\\
HigherHRNet\cite{cheng2019higherhrnet}  &67.6  &87.4  &72.6  &75.8  &68.1  &58.9\\
HRNet$\dagger$\cite{DBLP:journals/corr/abs-1902-09212}  &71.7  &89.8  &76.9  &79.6  &72.7  &61.5\\
DoubleEmbedding (Ours)  &68.8  &89.7  &73.4  &76.1  &69.5  &60.3\\
\hline
$\dagger$ indicates \quad top-down\quad methods
\end{tabular}
\end{center}
\end{table}
\setlength{\tabcolsep}{1.4pt}

Table 5 shows experimental results on CrowdPose. 
The proposed model achieves 68.8\% AP which outperforms the existing bottom-up methods.
but the performance is still lower than the state-of-the-art top-down method, HRNet which has intrinsic advantage in accuracy due to its processing flow. However, it narrows the accuracy gap between other bottom-up methods and top-down methods with less complexity. 
The performance on CrowdPose dataset indicates the robustness of our method in crowded scene.

\begin{figure}
  \centering
  \includegraphics[width=120mm]{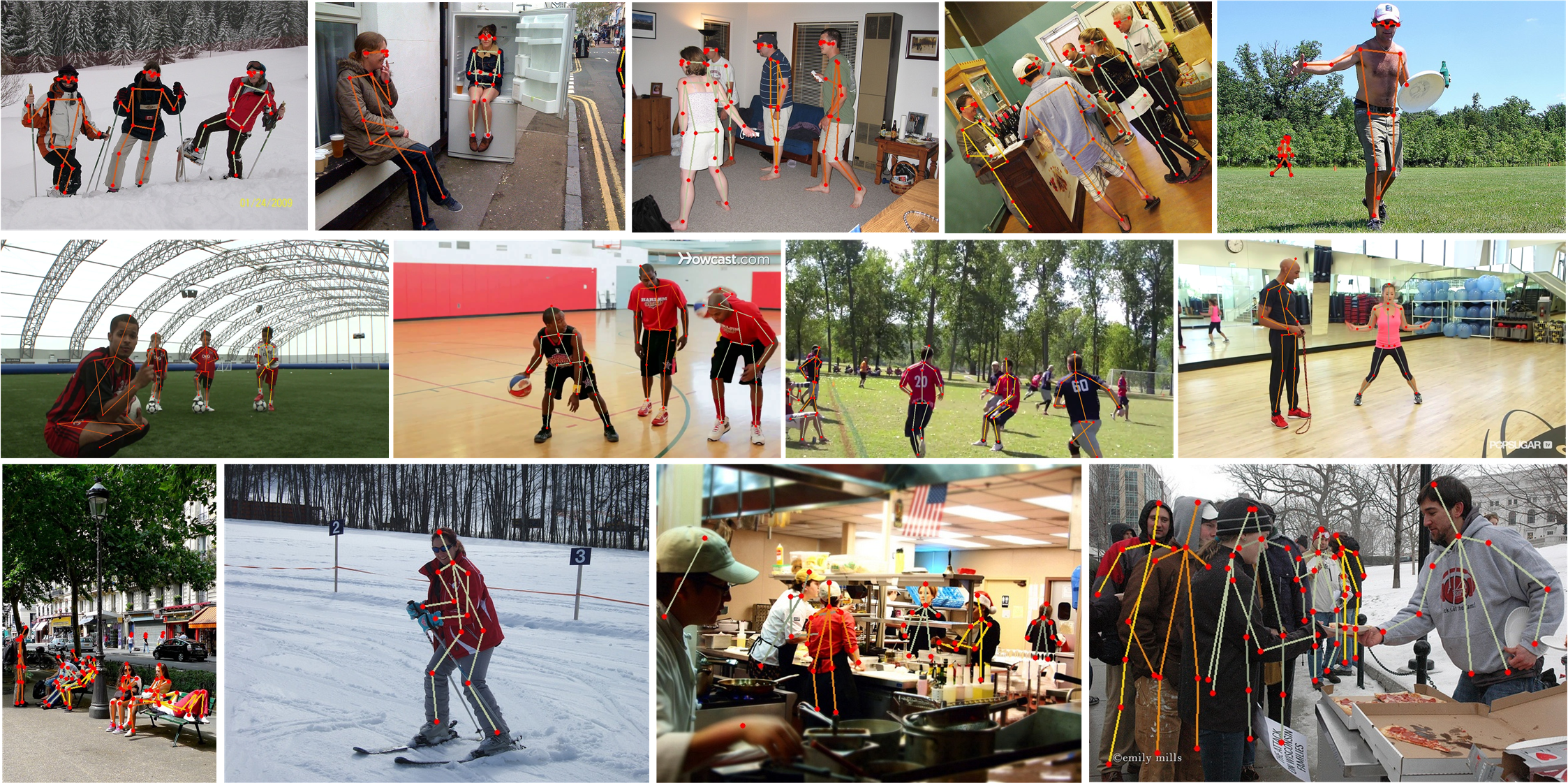} 
  \caption{
    Qualitative results on MSCOCO dataset (top), MPII dataset (middle) and CrowdPose dataset (bottom).
    }
  \label{fig:4}
  \end{figure}

Qualitative results on CrowdPose dataset are shown in the bottom row of Figure \ref{fig:4}. 
 The result verifies the effectiveness of our model in complex scenes, e.g., ambiguity pose and small-scale prediction (1st example), self-occluded (2nd example), cluttered background (3rd example) and person overlapping and crowded scene (4th example).
\newline
\section{Conclusion}
In this paper, we propose the Double Embedding (DE) method for multi-person pose estimation. Through Global Embedding (GE) and Local Embedding (LE), we achieve parallel implementation of person detection and joints detection, overcoming the intrinsic disadvantages of the conventional two-stage strategy on multi-person pose estimation. GE reduces the person instance detection problem to identifying a group of joints and LE connects and identifies the rest joints hierarchically. Based on LE, we design Mutual Refine Machine (MRM) to further enhance the performance for dealing with complex scenarios. We implement DE based on CNNs with end-to-end learning and inference. Experiments on three main benchmarks demonstrate the effectiveness of our model. DE achieves the competitive results among existing bottom-up methods and narrows the gap to the state-of-the-art top-down methods with less complexity.
\bibliographystyle{splncs}
\bibliography{egbib}

\begin{thebibliography}{10}

\bibitem{Ch2015P}
Chéron, G., Laptev, I., Schmid, C.:
\newblock P-cnn: Pose-based cnn features for action recognition.
\newblock (2015)

\bibitem{DBLP:journals/corr/abs-1811-08264}
Li, Y., Zhou, S., Huang, X., Xu, L., Ma, Z., Fang, H., Wang, Y., Lu, C.:
\newblock Transferable interactiveness prior for human-object interaction
  detection.
\newblock CoRR \textbf{abs/1811.08264} (2018)

\bibitem{Fang2018Pairwise}
Fang, H.S., Cao, J., Tai, Y.W., Lu, C.:
\newblock Pairwise body-part attention for recognizing human-object
  interactions.
\newblock (2018)

\bibitem{Qian2017Pose}
Qian, X., Fu, Y., Xiang, T., Wang, W., Qiu, J., Wu, Y., Jiang, Y.G., Xue, X.:
\newblock Pose-normalized image generation for person re-identification.
\newblock (2017)

\bibitem{Xiao2018Simple}
Xiao, B., Wu, H., Wei, Y.:
\newblock Simple baselines for human pose estimation and tracking.
\newblock (2018)

\bibitem{Sun2019Deep}
Sun, K., Xiao, B., Liu, D., Wang, J.:
\newblock Deep high-resolution representation learning for human pose
  estimation.
\newblock (2019)

\bibitem{Jingdong2020Deep}
Wang, J., Sun, K., Cheng, T., Jiang, B., Xiao, B.:
\newblock Deep high-resolution representation learning for visual recognition.
\newblock IEEE Transactions on Pattern Analysis and Machine Intelligence
  \textbf{PP} (2020)  1--1

\bibitem{DBLP:journals/corr/PapandreouZKTTB17}
Papandreou, G., Zhu, T., Kanazawa, N., Toshev, A., Tompson, J., Bregler, C.,
  Murphy, K.P.:
\newblock Towards accurate multi-person pose estimation in the wild.
\newblock CoRR \textbf{abs/1701.01779} (2017)

\bibitem{He2018Mask}
He, K., Georgia, G., Piotr, D., Ross, G.:
\newblock Mask r-cnn.
\newblock IEEE Transactions on Pattern Analysis \& Machine Intelligence (2018)
  1--1

\bibitem{Huang2017A}
Huang, S., Gong, M., Tao, D.:
\newblock A coarse-fine network for keypoint localization.
\newblock In: 2017 IEEE International Conference on Computer Vision (ICCV).
  (2017)

\bibitem{Fang2016RMPE}
Fang, H., Xie, S., Tai, Y., Lu, C.:
\newblock Rmpe: Regional multi-person pose estimation.
\newblock (2016)

\bibitem{Chen2017Cascaded}
Chen, Y., Wang, Z., Peng, Y., Zhang, Z., Yu, G., Sun, J.:
\newblock Cascaded pyramid network for multi-person pose estimation.
\newblock (2017)

\bibitem{Newell2016Stacked}
Newell, A., Yang, K., Deng, J.:
\newblock Stacked hourglass networks for human pose estimation.
\newblock (2016)

\bibitem{Pishchulin2016DeepCut}
Pishchulin, L., Insafutdinov, E., Tang, S., Andres, B., Andriluka, M., Gehler,
  P., Schiele, B.:
\newblock Deepcut: Joint subset partition and labeling for multi person pose
  estimation.
\newblock (2016)

\bibitem{Insafutdinov2016}
Insafutdinov, E., Pishchulin, L., Andres, B., Andriluka, M., Schiele, B.:
\newblock {DeeperCut}: {A} deeper, stronger, and faster multi-person pose
  estimation model.
\newblock (2016)

\bibitem{Iqbal2016Multi}
Iqbal, U., Gall, J.:
\newblock Multi-person pose estimation with local joint-to-person associations.
\newblock (2016)

\bibitem{osokin2018real}
Osokin, D.:
\newblock Real-time 2d multi-person pose estimation on cpu: Lightweight
  openpose.
\newblock arXiv preprint arXiv:1811.12004 (2018)

\bibitem{DBLP:journals/corr/NewellD16}
Newell, A., Deng, J.:
\newblock Associative embedding: End-to-end learning for joint detection and
  grouping.
\newblock CoRR \textbf{abs/1611.05424} (2016)

\bibitem{DBLP:journals/corr/abs-1803-08225}
Papandreou, G., Zhu, T., Chen, L., Gidaris, S., Tompson, J., Murphy, K.:
\newblock Personlab: Person pose estimation and instance segmentation with a
  bottom-up, part-based, geometric embedding model.
\newblock CoRR \textbf{abs/1803.08225} (2018)

\bibitem{2020HigherHRNet}
Cheng, B., Xiao, B., Wang, J., Shi, H., Zhang, L.:
\newblock Higherhrnet: Scale-aware representation learning for bottom-up human
  pose estimation.
\newblock In: 2020 IEEE/CVF Conference on Computer Vision and Pattern
  Recognition (CVPR). (2020)

\bibitem{Lin2014Microsoft}
Lin, T.Y., Maire, M., Belongie, S., Hays, J., Perona, P., Ramanan, D., Dollár,
  P., Zitnick, C.L.:
\newblock Microsoft coco: Common objects in context.
\newblock (2014)

\bibitem{Andriluka2014Human}
Andriluka, M., Pishchulin, L., Gehler, P., Schiele, B.:
\newblock Human pose estimation: New benchmark and state of the art analysis.
\newblock In: Computer Vision and Pattern Recognition (CVPR). (2014)

\bibitem{Li2018CrowdPose}
Li, J., Wang, C., Zhu, H., Mao, Y., Fang, H.S., Lu, C.:
\newblock Crowdpose: Efficient crowded scenes pose estimation and a new
  benchmark.
\newblock (2018)

\bibitem{Ren2015Faster}
Ren, S., He, K., Girshick, R., Sun, J.:
\newblock Faster r-cnn: Towards real-time object detection with region proposal
  networks.
\newblock IEEE Transactions on Pattern Analysis and Machine Intelligence (2015)

\bibitem{Lin2016Feature}
Lin, T.Y., Dollár, P., Girshick, R., He, K., Hariharan, B., Belongie, S.:
\newblock Feature pyramid networks for object detection.
\newblock (2016)

\bibitem{DBLP:journals/corr/abs-1803-06799}
Cheng, B., Wei, Y., Shi, H., Feris, R.S., Xiong, J., Huang, T.S.:
\newblock Revisiting {RCNN:} on awakening the classification power of faster
  {RCNN}.
\newblock CoRR \textbf{abs/1803.06799} (2018)

\bibitem{Cheng2018Decoupled}
Cheng, B., Wei, Y., Shi, H., Feris, R., Xiong, J., Huang, T.:
\newblock Decoupled classification refinement: Hard false positive suppression
  for object detection.
\newblock (2018)

\bibitem{Gkioxari2014Using}
Gkioxari, G., Hariharan, B., Girshick, R., Malik, J.:
\newblock Using k-poselets for detecting people and localizing their keypoints.
\newblock In: 2014 IEEE Conference on Computer Vision and Pattern Recognition
  (CVPR). (2014)

\bibitem{Min2011Articulated}
Min, S., Savarese, S.:
\newblock Articulated part-based model for joint object detection and pose
  estimation.
\newblock In: IEEE International Conference on Computer Vision, ICCV 2011,
  Barcelona, Spain, November 6-13, 2011. (2011)

\bibitem{7780880}
{Wei}, S., {Ramakrishna}, V., {Kanade}, T., {Sheikh}, Y.:
\newblock Convolutional pose machines.
\newblock In: 2016 IEEE Conference on Computer Vision and Pattern Recognition
  (CVPR). (2016)  4724--4732

\bibitem{fang2017rmpe}
Fang, H.S., Xie, S., Tai, Y.W., Lu, C.:
\newblock {RMPE}: Regional multi-person pose estimation.
\newblock In: ICCV. (2017)

\bibitem{Jaderberg2015Spatial}
Jaderberg, M., Simonyan, K., Zisserman, A.,  et~al.:
\newblock Spatial transformer networks.
\newblock In: Advances in neural information processing systems. (2015)
  2017--2025

\bibitem{DBLP:journals/corr/CaoSWS16}
Cao, Z., Simon, T., Wei, S., Sheikh, Y.:
\newblock Realtime multi-person 2d pose estimation using part affinity fields.
\newblock CoRR \textbf{abs/1611.08050} (2016)

\bibitem{DBLP:journals/corr/abs-1903-06593}
Kreiss, S., Bertoni, L., Alahi, A.:
\newblock Pifpaf: Composite fields for human pose estimation.
\newblock CoRR \textbf{abs/1903.06593} (2019)

\bibitem{DBLP:journals/corr/CarreiraAFM15}
Carreira, J., Agrawal, P., Fragkiadaki, K., Malik, J.:
\newblock Human pose estimation with iterative error feedback.
\newblock CoRR \textbf{abs/1507.06550} (2015)

\bibitem{Girshick2015Fast}
Girshick, R.:
\newblock Fast r-cnn.
\newblock Computer Science (2015)

\bibitem{nie2019singlestage}
Nie, X., Zhang, J., Yan, S., Feng, J.:
\newblock Single-stage multi-person pose machines (2019)

\bibitem{Kingma2014Adam}
Kingma, D., Ba, J.:
\newblock Adam: A method for stochastic optimization.
\newblock Computer Science (2014)

\bibitem{cheng2019higherhrnet}
Cheng, B., Xiao, B., Wang, J., Shi, H., Huang, T.S., Zhang, L.:
\newblock Higherhrnet: Scale-aware representation learning for bottom-up human
  pose estimation (2019)

\bibitem{DBLP:journals/corr/InsafutdinovAPT16}
Insafutdinov, E., Andriluka, M., Pishchulin, L., Tang, S., Levinkov, E.,
  Andres, B., Schiele, B.:
\newblock Articulated multi-person tracking in the wild.
\newblock CoRR \textbf{abs/1612.01465} (2016)

\bibitem{DBLP:journals/corr/abs-1804-07909}
Fieraru, M., Khoreva, A., Pishchulin, L., Schiele, B.:
\newblock Learning to refine human pose estimation.
\newblock CoRR \textbf{abs/1804.07909} (2018)

\bibitem{DBLP:journals/corr/abs-1902-09212}
Sun, K., Xiao, B., Liu, D., Wang, J.:
\newblock Deep high-resolution representation learning for human pose
  estimation.
\newblock CoRR \textbf{abs/1902.09212} (2019)

\end{thebibliography}
\end{document}